%% file: main.tex
\documentclass[10pt,twocolumn,letterpaper]{article}

\usepackage[pagenumbers]{iccv} 

\input{preamble}
\usepackage{booktabs}
\usepackage{color}
\usepackage{multirow}
\usepackage{arydshln}
\definecolor{iccvblue}{rgb}{0.21,0.49,0.74}
\usepackage[pagebackref,breaklinks,colorlinks,allcolors=iccvblue]{hyperref}

\robustify\cellcolor
\newcommand{\bgg}{\cellcolor[HTML]{B3E5FC}}
\newcommand{\bgo}{\cellcolor[HTML]{E3F6FF}}

\def\paperID{10222} 
\def\confName{ICCV}
\def\confYear{2025}

\title{Factorized Video Autoencoders for Efficient Generative Modelling}

\author{\begin{tabular}{cccc}
Mohammed Suhail$^{1}$ & Carlos Esteves$^{1}$ & Leonid Sigal$^{2,3,4}$ & Ameesh Makadia$^{1}$\\
 {\tt\small suhailmhd@google.com} & {\tt\small machc@google.com} &
 {\tt\small lsigal@cs.ubc.ca} & {\tt\small makadia@google.com}
\end{tabular}
\\
\begin{tabular}{lccr}
$^1$Google &
$^2$University of British Columbia &
$^3$Vector Institute for AI &
$^4$Canada CIFAR AI Chair 
\end{tabular}}

\begin{document}
\maketitle
\input{sec/0_abstract}    
\input{sec/1_intro}

\input{sec/2_related_work}
\input{sec/3_background}
\input{sec/4_method}
\input{sec/5_experiments}
{
    \small
    \bibliographystyle{ieeenat_fullname}
    \bibliography{main}
}

\input{sec/X_suppl}

\end{document}

%% file: preamble.tex
\newcommand{\fpshort}{4Plane}

%% file: sec/0_abstract.tex
\begin{abstract}
Latent variable generative models have emerged as powerful tools for generative tasks including image and video synthesis. These models are enabled by pretrained autoencoders that map high resolution data into a compressed lower dimensional latent space, where the generative models can subsequently be developed while requiring fewer computational resources. Despite their effectiveness, the direct application of latent variable models to higher dimensional domains such as videos continues to pose challenges for efficient training and inference. In this paper, we propose an autoencoder that projects volumetric data onto a four-plane factorized latent space that grows sublinearly with the input size, making it ideal for higher dimensional data like videos. The design of our factorized model supports straightforward adoption in a number of conditional generation tasks with latent diffusion models (LDMs), such as class-conditional generation, frame prediction, and video interpolation.
Our results show that the proposed four-plane latent space retains a rich representation needed for high-fidelity reconstructions despite the heavy compression, while simultaneously enabling LDMs to operate with significant improvements in speed and memory.
\end{abstract}

%% file: sec/1_intro.tex
\section{Introduction}
\label{sec:intro}

A defining trait of recent advances in image and video generation is that, as models grow more powerful, they increasingly push against the boundaries of current computational limits.  Despite their impressive generative capabilities, these models' vast resource demands hinder scalability and discourage widespread deployment.   Naturally, improving the efficiency of these generative models has become an active research concern~\cite{gupta2023walt,yu2024efficient,yu23cvpr,an2023latent}.

\begin{figure}[t!]
\includegraphics[width=.9\linewidth]{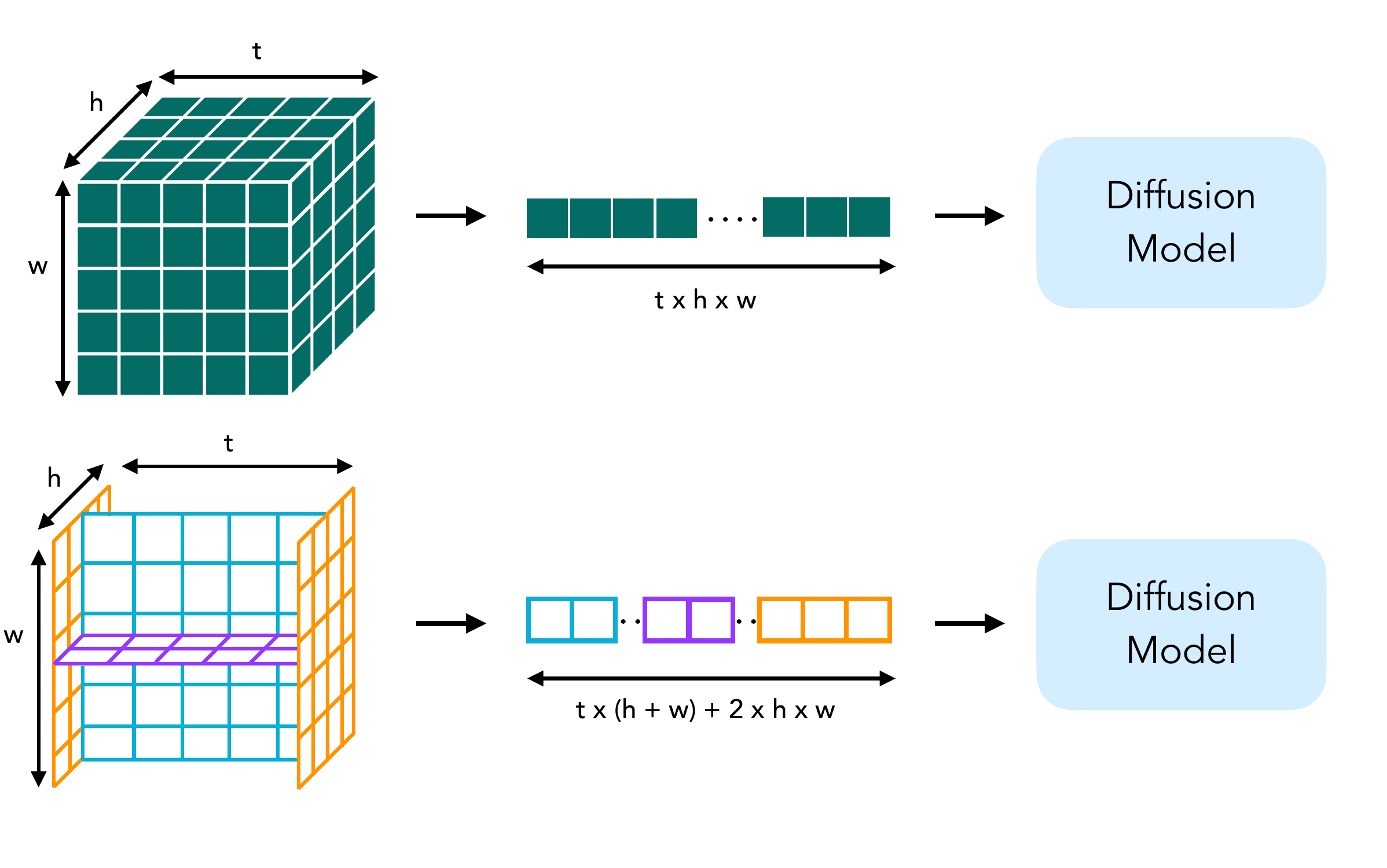}
\caption{\textbf{Factorized latent representation}. Traditional volumetric latents in diffusion models yield a sequence length of \( t \times h \times w \) (top row), which scales linearly with the input size and demands high computational resources. Our proposed factorized representation reduces sequence length to \( t \times (h + w) + 2 \times h \times w \) (bottom row), achieving a more compact latent space that scales sublinearly with input size, enabling faster, more efficient video generation without sacrificing quality.}
\label{fig:teaser}
\end{figure}

One effective strategy to make generative modeling computationally feasible is through latent modeling~\cite{blattmann2023stable,rombach2021high,blattmann2023videoldm,podell2023sdxl,dai2023emu,gupta2023walt,polyak2024movie,girdhar2023emu,an2023latent}. By compressing high-resolution visual data into a compact latent space, latent models significantly reduce the computational burden for generative models. However, in typical latent model autoencoders, the resulting embedding size still scales linearly with the original input size, so this compression offers only a limited benefit when deployed in very high dimensional domains, such as  videos~\cite{gupta2023walt,polyak2024movie} (see \Cref{fig:teaser}).

In this paper, we explore improving the efficiency of latent generative models through more aggressive reduction of the latent resolution. The central objective is to achieve this compression without sacrificing representation quality. To address this challenge, we propose a novel \textit{four-plane factorized latent autoencoder} that maps volumetric space-time signals onto a latent space through four axis-aligned planar projections. Since the orthogonal projections capture complementary features of the space-time volume, the original signal can still be reconstructed from this more compact latent space with high fidelity.  
We summarize the key attributes of our contribution below:

\begin{itemize}
\item \emph{Compact yet expressive factorization:} four-plane factorization significantly compresses volumetric latent embeddings, scaling sublinearly with the total input size (see~\cref{fig:teaser}). Despite this compression, it retains high-fidelity reconstructions.
\item \emph{Faster generation without sacrificing quality:} we demonstrate the factorized latent space is suitable for high quality generative modeling. A conventional transformer-based diffusion model trained to generate factorized latents is twice as fast compared to producing volumetric features~\cite{gupta2023walt}.
\item \emph{Versatility for image-conditioned tasks:} our experiments show how the four-plane structure seamlessly accommodates a variety of applications such as two-frame interpolation and future frame prediction.
\end{itemize}

Across a variety of tasks, our experiments suggest the proposed four-plane factorized autoencoder provides an efficient alternative for generative models that traditionally operate on volumetric latent spaces.

%% file: sec/2_related_work.tex
\section{Related work}
\label{sec:related_work}

\subsection{Diffusion models for video synthesis}
Denoising Diffusion Probabilistic Models (DDPM)~\cite{ho2020denoising} introduced a novel method for generating images by iteratively denoising a sequence of noisy images. This approach has been highly successful for both image~\cite{dhariwal2021diffusion,saharia2022photorealistic,podell2023sdxl,dai2023emu,hoogeboom2023simple} and video synthesis~\cite{ho2022video,Blattmann_2023_CVPR,ho2022imagen,wu2023tune,guo2023animatediff,blattmann2023stable}.

Of the more recent diffusion models developed for video generation, many operate on a volumetric spatiotemporal latent space. VDM~\cite{ho2022video} employs a 3D U-Net autoencoder architecture~\cite{cciccek20163d,ronneberger2015u} to learn this space. 
To address scalability for high resolution video generation, Imagen Video~\cite{ho2022imagen} extends VDM by introducing a cascade of models that essentially alternate temporal and spatial superresolution.  Lumiere~\cite{bar2024lumiere} introduced the STUNet architecture, which generates entire videos directly with improved temporal coherence.  VideoLDM~\cite{Blattmann_2023_CVPR} constructs a video model starting with pretrained image models and inserting temporal layers before fine-tuning on high-quality videos.

\subsection{Video tokenizers}
Many of the latent video diffusion models highlighted above rely on some form of video tokenization~\cite{gupta2023walt,zhao2025cv,yang2024cogvideox,agarwal2025cosmos,wang24omni} to compress high dimensional videos into a compact latent space.  The pioneering vector quantization approaches for image tokenization, for example VQ-VAE~\cite{van2017neural}, VQ-VAE2~\cite{razavi2019generating}, and VQGAN~\cite{esser2021taming}, can be applied to videos in a frame-by-frame manner.  MAGVIT~\cite{yu2023magvit} introduced a 3D-VQ autoencoder to quantize video data into spatio-temporal tokens, making it a powerful tool for a range of video generation tasks such a frame prediction, video inpainting, and frame interpolation. MAGVIT-v2~\cite{yu2023language} introduced significant advancements, including a lookup-free quantization method, which allows for an expanded vocabulary without compromising performance. Additionally, MAGVIT-v2 enables joint image and video modeling through a causal 3D CNN architecture. The MAGVIT-v2 tokenizer was used successfully in W.A.L.T~\cite{gupta2023walt} for photorealistic image and video generation.  
The early 3D encoder layers in our factorized architecture are based on MAGVIT-v2.


\subsection{Video frame interpolation}
Video frame interpolation~\cite{kiefhaber2024benchmarking,dong2023video} has distinct interpretations depending on the temporal distance between frames. In case with significant motion between frames, the challenging task can be addressed by generative models.  The application of our factorized latent representation to frame interpolation can be categorized as a two-frame conditioned diffusion model.  The first such effort to use a latent diffusion model was LDMVFI~\cite{danier2024ldmvfi}. In contrast, VIDIM~\cite{jain2024vidim} models in pixel space and generates the entire video at once improving temporal consistency. To improve quality, VIDIM employs a cascaded diffusion approach: it first generates a low-resolution video, followed by an upsampling diffusion model that refines the output to higher resolutions

\subsection{Tri-plane factorization}
Tri-plane representations, which factorize volumetric data into three orthogonal 2D planes, have been widely used as compact representations of 3D neural fields~\cite{shue20223cvpr,chen2022tensorf,fridovich2023k}.  Coupled with diffusion model architectures suited to planar representations, they have been used for a variety of applications, such as textured 3D model generation~\cite{wu2023sin3dm}, 3D neural field generation~\cite{shue20223cvpr}, and semantic scene generation~\cite{lee2024semcity}.  The same concept was applied to videos in PVDM~\cite{yu2023video}, where encoding videos to tri-plane latent features enables the use of a 2D U-Net architecture~\cite{dhariwal2021diffusion} for training diffusion models, bringing significant improvements in efficiency.  However, tri-plane approaches are still quite far behind their volumetric counterparts in generation quality, and the tri-plane representation cannot support different use cases. Both of these points are addressed by our distinctive four-plane factorization.

%% file: sec/3_background.tex
\begin{figure*}[t!]
\centering
\includegraphics[width=\linewidth]{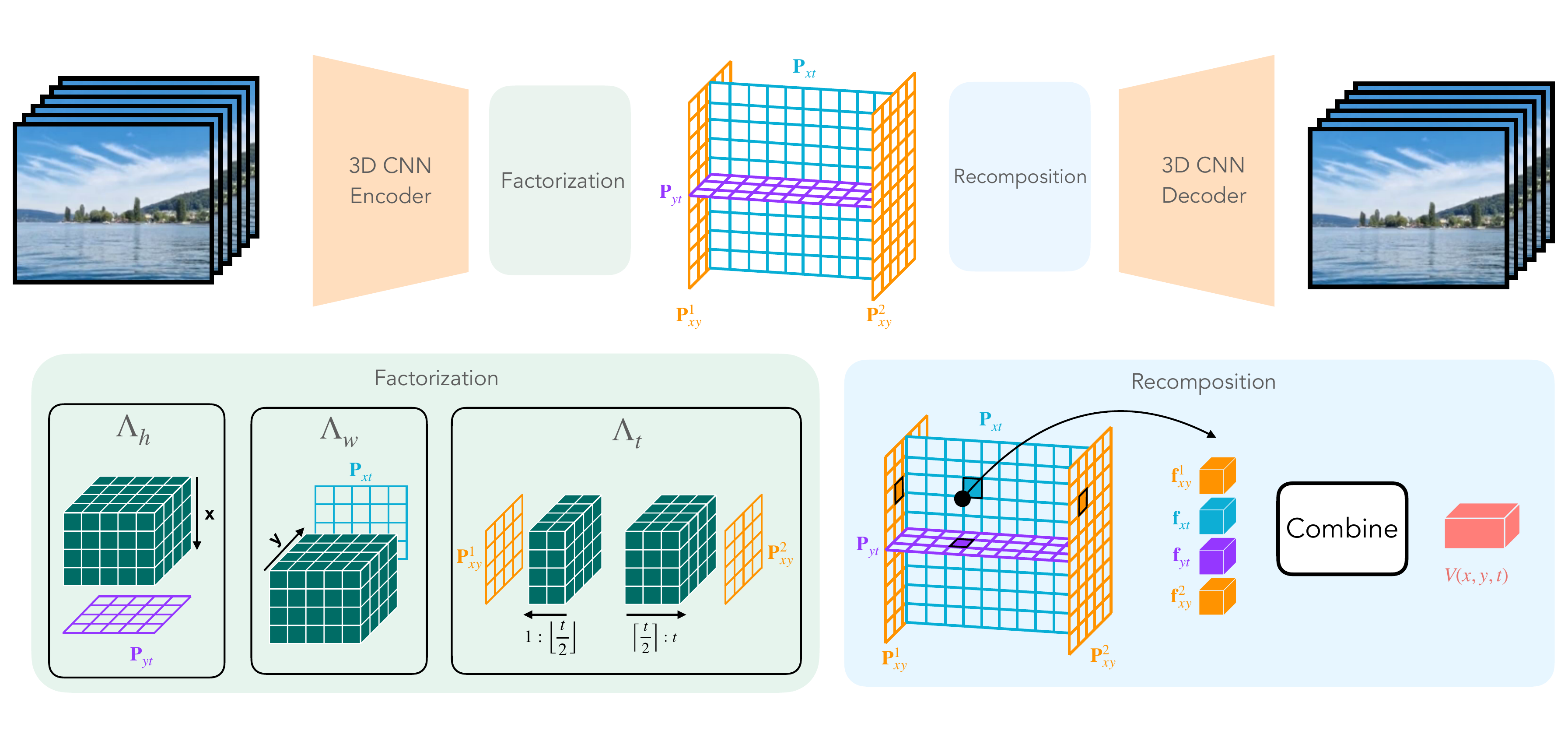}
\caption{\textbf{Model overview.} The autoencoder first maps the input video into a volumetric latent representation through \(3D\) convolutional architecture, which is then factorized into four planes. Temporal planes are created by mean pooling along the height and width dimensions, capturing time-varying features. Spatial planes are obtained by splitting along the time axis and independently averaging along this dimension, focusing on spatial consistency (highlighted in green). During decoding, the four planes are mapped back into a volume: for each spatial-temporal location, features from the corresponding four planes (highlighted in blue) are concatenated to reconstruct the full volumetric features. These combined features are passed through a decoder to produce the final output video.}
\label{fig:model}
\end{figure*}

%% file: sec/4_method.tex
\section{Factorized video latent representations}

In latent video diffusion models, a key component is the autoencoder, which compresses the input video data into a compact latent representation. 
To achieve this, prior works~\cite{gupta2023walt,he2022lvdm} employ a 3D convolutional architecture that encodes the 3D video volume $\mathbf{X} \in \mathbb{R}^{T \times H \times W \times C}$ into a feature volume $\mathbf{Z} \in \mathbb{R}^{t \times h \times w \times c}$, where $t = \frac{T}{f_{t}}$, $h = \frac{H}{f_s}$, $w = \frac{W}{f_s}$. Here $f_t$ and $f_s$ are the temporal and spatial downsampling factors. 
The channel dimension $c$ is typically expanded ({\em e.g.}, $c = 8$ is a common choice that balances autoencoder reconstruction and diffusion performance).

While this compression does offer significant reduction in the spatial and temporal resolution, the total size ($t \times h \times w$) still scales linearly with the input size. For computationally expensive generative models such as transformer~\cite{NIPS2017_3f5ee243}-based diffusion models, this sequence length would still pose a challenge.  Improving the efficiency of transformer-based models can either be achieved by addressing the design of the model itself ({\em e.g.}, sub-quadratic attention mechanisms). or by decreasing the sequence length. In this work we explore the latter by introducing a four-plane factorized autoencoder that we describe in the following section.



\subsection{Four plane factorization}
\label{sec:four_plane_factorization}
Our approach provides a direct and effective solution to reducing the cubic complexity of volumetric spatiotemporal latent spaces. By decomposing the 3D feature volume into four complementary planes, our method captures rich spatial and temporal structures while significantly improving efficiency. This streamlined representation not only accelerates training and inference but also reduces memory overhead without compromising reconstruction quality. Beyond efficiency gains, the four-plane factorization introduces a versatile framework adaptable to a wide range of video generation tasks, including class-conditional generation, frame extrapolation, and video interpolation.

\subsubsection{Factorization.}
\label{sec:factorization}
Given an input video \(\mathbf{X} \in \mathbb{R}^{T \times H \times W \times 3}\), our encoder network first converts it into a feature volume $\mathbf{Z} \in \mathbb{R}^{t \times h \times w \times c}$ using a causal 3D convolution architecture similar to the one introduced in MAGVIT-v2~\cite{yu2024magvitv2}. The feature volume is then factorized into four planes along three directions: two spatial planes, \(\mathbf{P}_{xy}^{1}, \mathbf{P}_{xy}^{2} \in \mathbb{R}^{h \times w \times c}\), aligned along the \(xy\)-dimension, and two spatiotemporal planes, \(\mathbf{P}_{xt} \in \mathbb{R}^{t \times h \times c}\) and \(\mathbf{P}_{yt} \in \mathbb{R}^{t \times w \times c}\), aligned along the \(xt\) and \(yt\) dimensions, respectively. 
The two spatio-temporal planes $\mathbf{P}_{xt}$ and $\mathbf{P}_{yt}$ are obtained by collapsing the height and width dimensions, respectively:
\begin{align}
    \mathbf{P}_{xt} &= \Lambda_{w}(\mathbf{Z}) \in \mathbb{R}^{t \times h \times c} \\
    \mathbf{P}_{yt} &= \Lambda_{h}(\mathbf{Z}) \in \mathbb{R}^{t \times w \times c}
\end{align}
where $\Lambda_{h}$ and $\Lambda_{w}$ performs pooling or dimensionality reduction operation along the required dimensions.

The spatial planes \(\mathbf{P}_{xy}^{1}\) and \(\mathbf{P}_{xy}^{2}\) contain the temporally aggregated features across frames, capturing the spatial structure and background information in the video.  
We adapt our factorization approach based on the application. For class-conditional generation and frame prediction, we split the latent feature volume \(\mathbf{Z} \in \mathbb{R}^{t \times h \times w \times c}\) into two segments along the temporal dimension, then aggregate each segment over \(T\), yielding:
\begin{equation}
\mathbf{P}_{xy}^{1} = \Lambda_{t}(\mathbf{Z}_{1:\lfloor\frac{t}{2}\rfloor}), \quad \mathbf{P}_{xy}^{2} = \Lambda_{t}(\mathbf{Z}_{\lceil\frac{t}{2}\rceil:t}),
\end{equation}
where \(\Lambda_{t}\) is an aggregation function similar to $\Lambda_{h,w}$, \(\lfloor \cdot  \rfloor\) and \(\lceil \cdot  \rceil\) represents the floor and ceil function respectively.

For the interpolation task only the first and last frames are available at inference and therefore the spatial plane cannot contain information from other timesteps. To address this, we obtain the spatial planes by encoding only the first and last frames using our encoder. Specifically, we set:
\begin{equation}
\mathbf{P}_{xy}^{1} = E(\mathbf{X}_{0}), \quad \mathbf{P}_{xy}^{2} = E(\mathbf{X}_{T}),
\end{equation}
where \(E\) denotes our video encoder and \(\mathbf{X}_{0}\) and \(\mathbf{X}_{T}\) are the boundary frames. Since our model uses a 3D CNN with causal temporal padding, it can naturally encode images without requiring additional modifications. This approach effectively incorporates key frame information into the spatial planes, enhancing the model’s interpolation accuracy.

\subsubsection{Recomposition}
\label{sec:recomposition}
Given the four latent planes \(\mathbf{P}_{xy}^{1}, \mathbf{P}_{xy}^{2}, \mathbf{P}_{xt},\) and \(\mathbf{P}_{yt}\), the decoder reconstructs the input video by first reconstituting these planes back into a 3D feature volume. To utilize existing 3D convolutional architectures, we construct an intermediate volume \(\mathbf{V} \in \mathbb{R}^{t \times h \times w \times c}\) by back-projecting features from each plane onto corresponding locations within the target volume dimensions \((t, h, w)\). 

For any spatial-temporal location \((x, y, t)\) in the volume, we extract features from each of the planes by projecting onto their respective dimensions (depicted in the blue box in~\Cref{fig:model}). Specifically:
\begin{align*}
    \mathbf{f}_{xy}^{1} = \mathbf{P}_{xy}^{1}(x, y), &\quad \mathbf{f}_{xy}^{2} = \mathbf{P}_{xy}^{2}(x, y) \\
    \mathbf{f}_{xt} = \mathbf{P}_{xt}(x, t), & \quad \mathbf{f}_{yt} = \mathbf{P}_{yt}(y, t).
\end{align*}

\noindent
Here \(\mathbf{f}_{xy}^{1}\),  \(\mathbf{f}_{xy}^{2}\), \(\mathbf{f}_{xt}\), and \(\mathbf{f}_{yt}\) will contain features queried from their respective planes, using the corresponding spatial or temporal coordinates. These features are then combined using an operation such as element-wise addition or concatenation, yielding:
\[
\mathbf{V}(x, y, t) = \mathrm{Combine}(\mathbf{f}^{1}_{xy}, \mathbf{f}^{2}_{xy}, \mathbf{f}_{xt}, \mathbf{f}_{yt}),
\]
where \(\mathrm{Combine}\) represents the recomposition  operation, e.g., summation or concatenation, along the channel dimension.

The reconstructed feature volume \(\mathbf{V}\) is then fed into a decoder with a structure similar to MAGVIT-v2, which progressively upsamples the features and applies 3D convolutions to generate the final video \(\hat{\mathbf{X}} \in \mathbb{R}^{T \times H \times W \times 3}\).

\subsection{Generative modeling with factorized latents}
\label{secdiffusion}
With a trained factorized latent model, we obtain a compact, efficient representation of input video data, suitable for training generative models.  In our experiments we utilize established techniques for transformer-based latent diffusion models.

Latent diffusion models (LDMs) gradually transform the latent representation of data into noise in a forward diffusion process, then reverse this transformation to generate new samples. Given an initial latent sample \(\mathbf{x}_0\), the forward process generates a sequence of increasingly noisy latents \(\{\mathbf{z}_t\}_{t=1}^{T}\) as \(\mathbf{z}_t = \sqrt{\alpha_t} \mathbf{z}_{t-1} + \sqrt{1 - \alpha_t} \mathbf{\epsilon}, \quad \mathbf{x}_0 \approx D(\mathbf{z}_0) \)
where $\alpha_t$ controls the noise schedule and $D$ decodes the final latent back to data space. The reverse process, defined by $p_\theta(\mathbf{z}_{t-1} | \mathbf{z}_t) = \mathcal{N}(\mathbf{z}_{t-1}; \mu_\theta(\mathbf{z}_t, t), \sigma_t^2 \mathbf{I})$, seeks to denoise the latent variables and reconstruct the original data distribution. This denoising is learned by minimizing a variational lower bound, often simplified into practical objectives~\cite{ho2020denoising}. 
Here, we adopt the v-parameterization, following recent diffusion improvements~\cite{salimans2022progressive}.


To train a LDM on our factorized representation we use a transformer architecture.  We create a 1D sequence by flattening the four planes—\(\mathbf{P}_{xt}\), \(\mathbf{P}_{yt}\), \(\mathbf{P}^{1}_{xy}\), and \(\mathbf{P}^2_{xy}\)—and concatenating them along the sequence length dimension. This results in a sequence length of \(h \times t + w \times t + 2 \times h \times w\) (as shown in~\Cref{fig:teaser}).



%% file: sec/5_experiments.tex
\section{Experiments}
The focus of our experiments is to show that the four-plane factorized model can generally and seamlessly replace volumetric latent spaces when modeling videos.  To understand the model's representation capability, we will quantify the compression versus reconstruction tradeoff (\cref{sec:exp-reconstruction}).  To demonstrate its widespread applicability, we deploy the factorized latent space in varied generative tasks such as class-conditional video generation (\cref{sec:exp-cc}), future frames prediction (\cref{sec:exp-prediction}), and two-frame interpolation (\cref{sec:exp-interp}). Notably, the factorized autoencoder design is used without modification across experiments, as only the training data changes to ensure fair comparisons for each task.

\subsection{Autoencoder reconstruction}
\label{sec:exp-reconstruction}
One primary consideration of the proposed factorization is if the considerable latent compression comes at a cost in representation quality.  For this, we compare the reconstructions from a baseline volumetric autoencoder against our factorized approach. The factorized encoder and decoder share the same architecture as the volumetric baseline, differing only in the factorization and recomposition steps. Specifically, to obtain the planes from the latent feature volume, we use average pooling as the factorization operations $\Lambda_{h}$ , $\Lambda_{w}$ and $\Lambda_t$.  To recompose the planes into feature volume, we use the concatenation operation as $\mathrm{Combine}$. We discuss further architecture and training details in the apendix.

We chose the W.A.L.T.~\cite{gupta2023walt} autoencoder as our volumetric baseline model as it has shown state of the art performance on multiple benchmark tasks such as class-conditional video generation and frame extrapolation.  Furthermore we were able to reproduce the model in terms of similar datasets and performance, allowing us to evaluate it in new settings as well.

We trained the volumetric and our four-plane factorized model on the Kinetics-600 (K600)~\cite{carreira2018short} dataset, consisting of nearly 400,000 video clips, covering around 600 action classes and exhibiting a wide range of human activities.  ~\cref{table:reconstruction} shows the reconstruction quality for \(128 \times 128\) resolution videos is nearly identical for both the volumetric and factorized autoencoders, while the latent size (sequence length) is nearly halved for the factorized. 
We also train a \(256 \times 256\) model by attaching an extra layer to the encoder and decoder architecture (see appendix for details), leaving the latent size unchanged and observe comparable performance for the volumetric and fourplane autoencoder.

Whereas W.A.L.T. adopts a continuous autoencoder framework, some recent works for latent video generation have opted for a variational autoencoder (VAE~\cite{kingma2013auto}) design~\cite{wang24omni,yang2024cogvideox,zhao2025cv}. Our factorized proposal is not predicated on this decision, so we would expect similar conclusions for our approach in the VAE setting. To validate this, we conduct two additional experiments. First, we modify W.A.L.T. autoencoder to incorporate a VAE decoder and construct a corresponding factorized-VAE variant. Second, we adopt WF-VAE~\cite{li2024wf} from OpenSoraPlan~\cite{lin2024open}, training the VAE using our factorized representation and comparing its performance against the baseline. For experiments with the W.A.L.T. autoencoder, we use a latent dimensionality of 8, while for WF-VAE, we set the latent dimensionality to 4. ~\Cref{table:reconstruction} confirms the relative performance holds across both VAE and AE settings.  
Since W.A.L.T. (AE) is the established baseline, our subsequent evaluations are all performed in the AE setting.


\begin{table}[]
\centering
\resizebox{\columnwidth}{!}{
\begin{tabular}{@{}llcccc@{}}
\toprule \toprule
          Res.           & Method        & PSNR$\uparrow$   & SSIM$\uparrow$  & LPIPS$\downarrow$ & Seq.Len \\ \midrule
\multirow{4}{*}{128} & Volumetric        & 27.64 & 0.85 & 0.049 & 1280    \\
                     & \fpshort          & 27.11 & 0.82 & 0.051 & 672     \\
                     & Volumetric-VAE     & 27.32 & 0.84 & 0.053 & 1280    \\
                     & \fpshort-VAE      & 27.03 & 0.82 & 0.055 & 672     \\ 
                     \midrule
\multirow{4}{*}{256} & Volumetric        & 26.27 & 0.79 & 0.089       & 1280    \\
                     & \fpshort          & 25.67 & 0.77 & 0.104 & 672     \\ 
                     & WF-VAE             & 27.86 & 0.83 & 0.064 & 1280    \\
                     & 4Plane-WF-VAE      & 26.98 & 0.81 & 0.073 & 672     \\ 
                     \bottomrule \bottomrule
\end{tabular}
}
\caption{\textbf{Video reconstruction.}
We show reconstruction metrics for our four-plane model and the volumetric baseline (W.A.L.T.~\cite{gupta2023walt}) on the Kinetics-600 test set, alongside the corresponding sequence length induced by each latent representation. We evaluate tokenizers trained using both autoencoder and VAE frameworks.}
\label{table:reconstruction}
\end{table}


\begin{table*}[t!]
\centering
\begin{tabular}{@{}lc@{\extracolsep{\tabcolsep}}c@{\extracolsep{3\tabcolsep}}c@{\extracolsep{3\tabcolsep}}c@{\extracolsep{\tabcolsep}}c@{}}
\toprule
\toprule
                                         & \multicolumn{2}{r@{\extracolsep{3\tabcolsep}}}{Class Conditional Generation (FVD $\downarrow$)} & Frame Prediction (FVD $\downarrow$) & \multicolumn{1}{@{}c@{\extracolsep{\tabcolsep}}}{\multirow{2}{*}{Params}} & \multicolumn{1}{@{}c@{\extracolsep{\tabcolsep}}}{\multirow{2}{*}{Steps}} \\
\cmidrule(l){2-3} \cmidrule{4-4}                                         
                                         & \small{UCF-101 (128x128)}                    & \small{UCF-101 (256x256)}                   & \small{Kinetics-600 (128x128)}            &        &       \\ \midrule
Video Diffusion~\cite{ho2022video}       & -                          & -                         & 16.2                   & 1.1B   & 256   \\
RIN ~\cite{jabri2022scalable}            & -                          & -                         & 10.7                   & 411M   & 1000  \\
Phenaki~\cite{villegas2022phenaki}       & -                          & -                         & 36.4                   & 227M   & 1024  \\
MAGVIT~\cite{yu2023magvit}               & 76                         & -                         & 9.9                    & 306M   & 48    \\
PVDM~\cite{yu23cvpr}                     & -                          & 399.4                     & -                      &  -     & 400   \\
MAGVIT-v2~\cite{yu2024magvitv2}          & 58                         & -                         & 4.3                    & 307M   & 24    \\
HVDM~\cite{kim2024hybrid}                & -                          & 303.1                     & -                      & 63M    & 100   \\
WALT~\cite{gupta2023walt}                & 46                         & -                         & \bgg 3.3                    & 313M   & 50    \\
WALT*                                    & \bgo 39                         & \bgo 84.68                     & \bgo 5.7                    & 214M   & 50    \\ \midrule
\fpshort~(Ours)                                     & \bgg 38                         & \bgg 58.27                     & 8.6                    & 214M   & 50    \\ \bottomrule \bottomrule
\end{tabular}
\caption{\textbf{Class-conditional generation on UCF and frame prediction on Kinetics-600.} WALT* represents our re-training and re-evaluation of the WALT baseline. UCF-128 and UCF-256 refer to exeperiments at 128 x 128 and 256 x 256 resolutios respectively. Our method achieves competitive performance with WALT on the UCF-128 task and performs slightly lower on the K600 frame prediction task, showcasing efficient performance across both datasets. On UCF-256 our model outperforms the baselines. }
\label{table:ucf_generation}
\end{table*}

\subsection{Class-conditional generation}
\label{sec:exp-cc}
We evaluate the four-plane factorized latent space in a variety of generative settings, starting with class-conditional generation.  Our evaluation is on the same two autoencoders described in the previous section, trained on K600 for 17 frame video reconstruction, at \(128 \times 128\) and \(256 \times 256\) resolutions.  For generation, we train a transformer-based diffusion model to generate factorized latent embeddings, following \Cref{secdiffusion}. The diffusion model is trained on  the UCF-101 dataset~\cite{soomro2012ucf101}, which comprises 9,537 videos spanning 101 action categories, offering a diverse set of motion dynamics.  To evaluate the quality of generated videos, we use the Fréchet Video Distance (FVD)~\cite{unterthiner2019fvd} as our primary metric. FVD measures the similarity between the distributions of generated and real videos, assessing both spatial realism and temporal coherence.

\begin{figure}[]
\centering
\includegraphics[width=\linewidth]{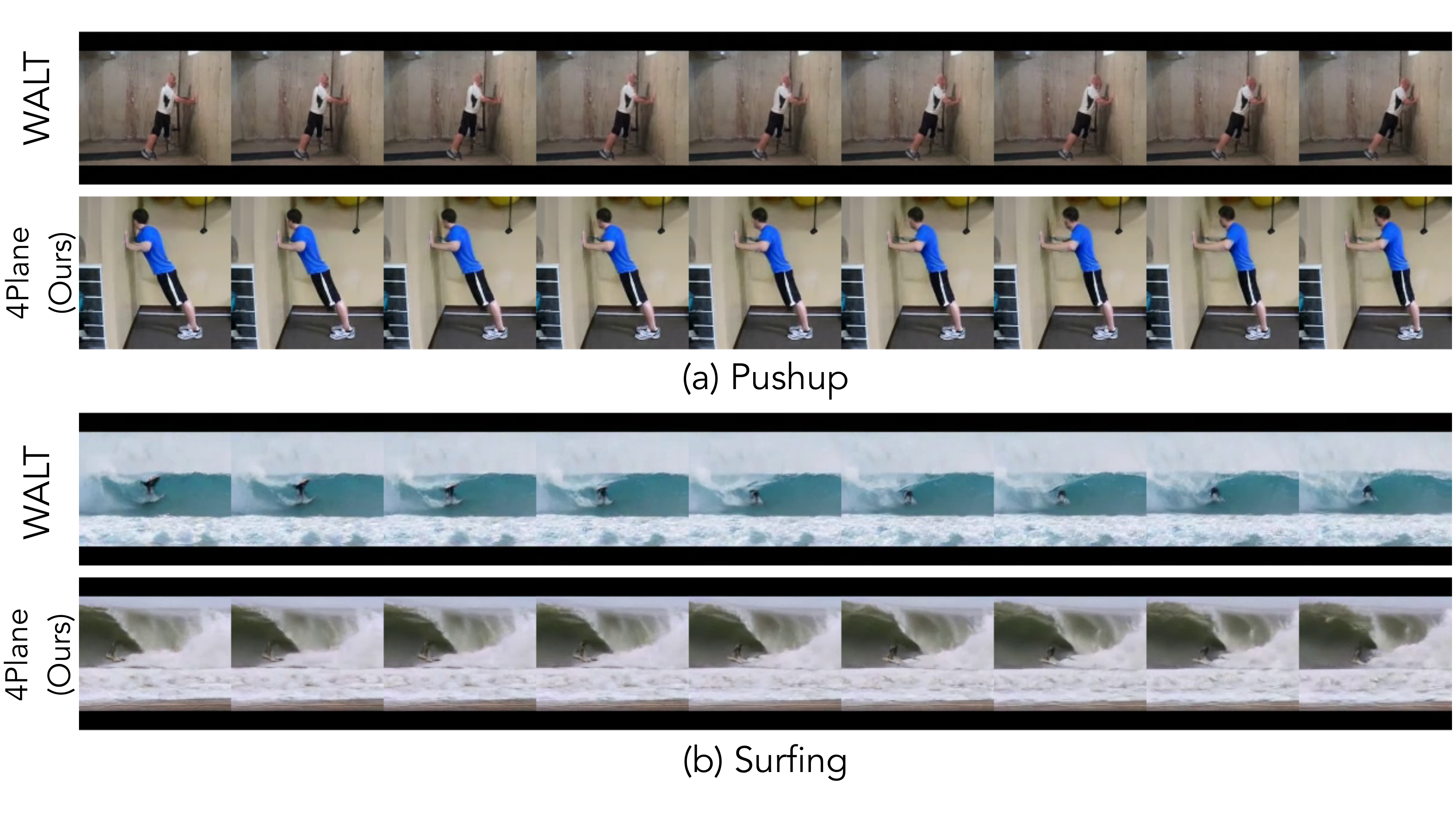}
\label{fig:ucf_results}
\caption{\textbf{Class-conditional generation results on the UCF dataset.} We show every other frame of the 17-frame generated videos from the \(128 \times 128\) models. The temporal continuity and overall frame quality of our factorized model is comparable to the volumetric W.A.L.T. generations.  
}
\end{figure}

 

\begin{table*}[t!]
\centering
\begin{tabular}{@{}llc@{\extracolsep{1.5\tabcolsep}}c@{\extracolsep{1.5\tabcolsep}}c@{\extracolsep{1.5\tabcolsep}}c@{\extracolsep{4\tabcolsep}}c@{\extracolsep{1.5\tabcolsep}}c@{\extracolsep{1.5\tabcolsep}}c@{\extracolsep{1.5\tabcolsep}}c@{}}
\toprule
\toprule
& & \multicolumn{4}{c}{Davis-7}              & \multicolumn{4}{c}{UCF-7}                \\ 
        \cmidrule(l){3-6} \cmidrule(l){7-10} 
 Type                                            & Method & PSNR$\uparrow$  & SSIM$\uparrow$   & LPIPS$\downarrow$    & FVD$\downarrow$    & PSNR$\uparrow$   & SSIM$\uparrow$   & LPIPS$\downarrow$     & FVD$\downarrow$    \\ \midrule
\multicolumn{1}{@{}l}{\multirow{3}{*}{Flow Based}}  & AMT~\cite{li2023amt}    & 21.09 & 0.544  & 0.254    & 234.5  & 26.06  & 0.813  & 0.144   & 344.5  \\
\multicolumn{1}{l}{}                             & RIFE~\cite{huang2022real}   & 20.48 & 0.511  & 0.258    & 240.0  & 25.73  & 0.804  & 0.135   & 323.8  \\
\multicolumn{1}{l}{}                             & FILM~\cite{reda2022film}   & 20.71 & 0.528  & 0.270    & 214.8  & 25.90  & 0.811  & 0.137   & 328.2  \\ 
\cdashline{1-10}
\multicolumn{1}{@{}l}{\multirow{3}{*}{Diffusion}}   & LDMVFI~\cite{danier2024ldmvfi} & 19.98 & 0.479  & 0.276    & 245.0  & 25.57  & 0.800  & 0.135   & 316.3  \\
\multicolumn{1}{l}{}                             & VIDIM~\cite{jain2024video}  & 19.62 & 0.470  & 0.257    & 199.3  & 24.07  & 0.781  & 0.149   & 278.0    \\ \cmidrule(l){2-10}
\multicolumn{1}{l}{}                             & \fpshort~(Ours)   & 19.47 & 0.446  & 0.256  & 156.1  & 24.00  & 0.769  & 0.141   & 216.9 \\ \bottomrule
\bottomrule
\end{tabular}
\caption{\textbf{Video interpolation results on DAVIS-7 and UCF-7}. Our method is compared against several video interpolation baselines, assessing both reconstruction and generative metrics, across all 7 interpolated frames.  See the text for additional discussion.}
\label{table:interpolation}
\end{table*}

\subsubsection{Diffusion training details}
\label{subsubsec:training}
For a fair comparison, we use a network architecture identical to W.A.L.T. Despite the architecture similarity, W.A.L.T. requires \(1.6 \times 10^{12}\) FLOPs whereas our model only uses \(8.5 \times 10^{11}\) due to shorter sequence lengths . We use a self-conditioning~\cite{chen2022analog} rate of \(0.9\), AdaLN-LoRA~\cite{gupta2023walt} with \(r=2\) as the conditioning mechanism and zero terminal SNR~\cite{lin2024common} to avoid mismatch between training and inference arising from non-zero signal-to-noise ratio at the final time in noise schedules.  We additionally use query-key normalization in the transformer to stabilize training. Our model is trained with a batch size of \(256\) using an Adam optimizer with a base learning rate of \(5 \times 10^{-4}\) with a linear warmup and cosine decay.

\subsubsection{Analysis}
\label{subsec:analysis}
At a resolution of \(128 \times 128\), our factorized model outperforms most prior works and performs comparably to W.A.L.T.~\cite{gupta2023walt} (see  ~\cref{table:ucf_generation}). Furthermore, our model achieves a higher Inception Score of 92.21, compared to 90.95 reported by W.A.L.T. Additionally, the smaller sequence length for our method makes diffusion training and inference nearly 2x faster compared to W.A.L.T. -- under identical training architecture and computational resources, our model processes each training iteration in just 380 ms compared to 750 milliseconds for W.A.L.T.  We show a detailed timing analysis in the appendix.  A comparison of qualitative results are provided in \cref{fig:ucf_results}.

At \(256 \times 256\) resolution, our factorization enables a diffusion model that outperforms prior baselines, despite the significant compression.  Recall that the factorized sequence length for our \(256 \times 256\) resolution model is the same as for \(128 \times 128\), resulting in 4x more compression. Nonetheless, the generation quality decays relatively much less. Moreover, the comparison to WALT* indicates that a shortened sequence that retains critical spatiotemporal information can in fact reduce the modeling burden on the denoiser network, thereby improving generation quality. We show qualitative comparison in the appendix.

Compared to the tri-plane factorization approaches PVDM~\cite{yu23cvpr} and HVDM~\cite{kim2024hybrid}, the four-plane factorization yields substantial improvements, validating the critical design choices of our model. An additional tri-plane ablation is provided in the appendix. Another concern is that the tri-plane structure introduces information mixing in the spatial planes, making the representation unsuitable for different tasks like frame prediction, which we evaluate next.

\begin{figure*}[t!]
\centering
\includegraphics[width=.85\linewidth]{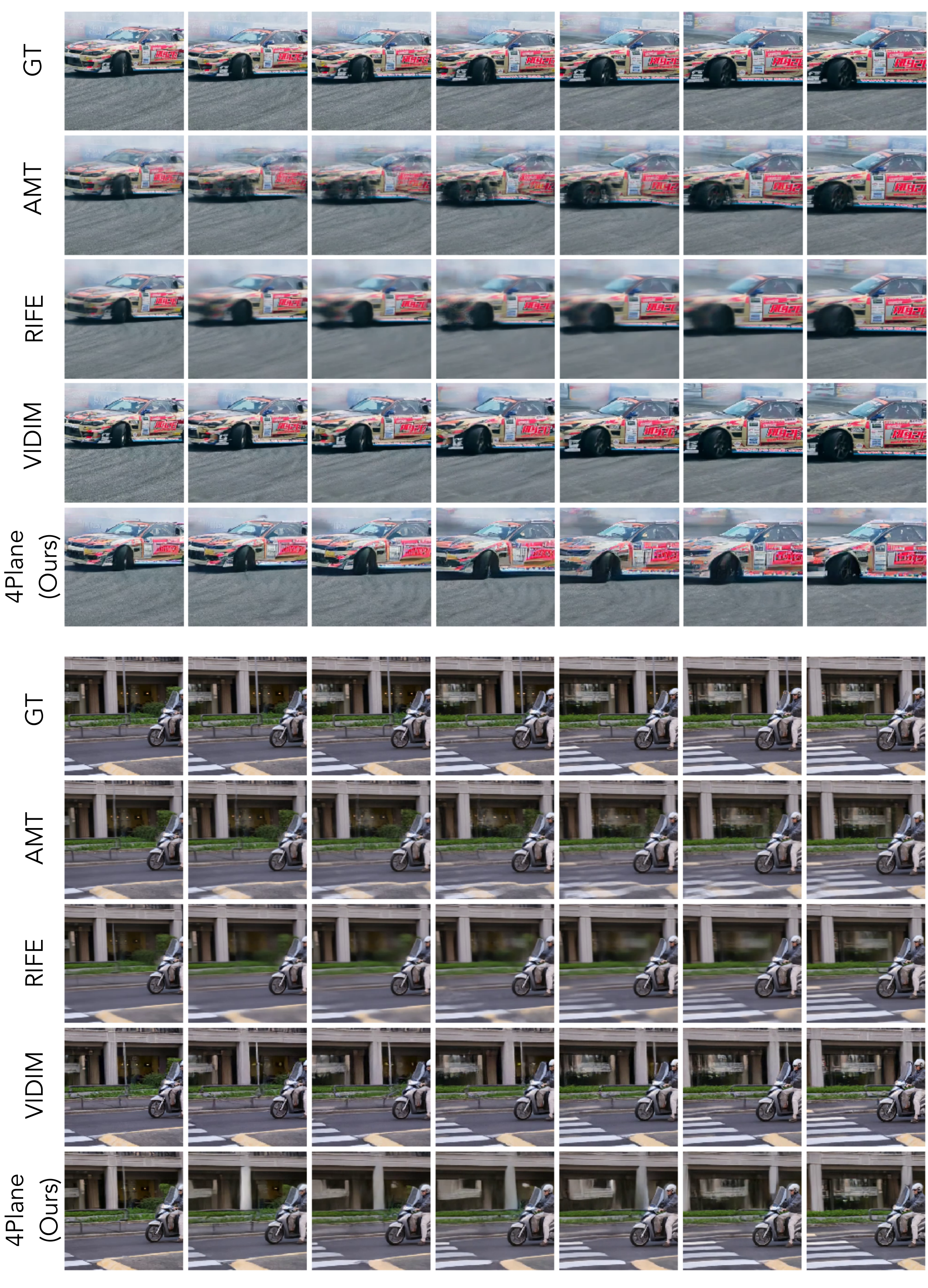}
\caption{\textbf{Interpolation results.} We show the \(7\) interpolated frames for two scenes from the DAVIS-7~\cite{jain2024vidim} dataset, our method generates realistic videos with sharp, detailed frames, achieving quality comparable to VIDIM~\cite{jain2024vidim}.}
\label{fig:interpolation}
\end{figure*}

\subsection{Future frame prediction}
\label{sec:exp-prediction}
For the future frame predicition task, we reuse the autoencoder trained for class-conditional generation (\Cref{sec:exp-cc}). The denoiser network architecture and training procedure remain consistent with the details provided in~\cref{subsubsec:training}, with the key difference that it is trained on K600 to align with the benchmark setting.. The diffusion model uses the first spatial plane $\mathbf{P}_{xy}^1$ as a conditioning sequence and learns to generate the remaining three planes $\mathbf{P}_{xy}^2$, $\mathbf{P}_{xt}$, and $\mathbf{P}_{yt}$. By leveraging a causal encoder, this setup mirrors W.A.L.T.'s frame prediction approach, where the model conditions on two latent frames. 
The frame prediction results in  ~\cref{table:ucf_generation} indicate our model outperforms most prior works and is comparable to WALT* while being significantly faster (see timing analysis in appendix). 

\subsection{Video interpolation}
\label{sec:exp-interp}
Video interpolation techniques generate intermediate frames between given keyframes, and are essential for applications requiring fluid motion reconstruction, such as frame-rate upsampling and video inpainting. In this experiment, we leverage our factorized latent representation to train the diffusion model to generate the spatio-temporal plane latents, $\mathbf{P}_{xt}$ and $\mathbf{P}_{yt}$, conditioned on the spatial plane latents, $\mathbf{P}_{xy}^1$ and $\mathbf{P}_{xy}^2$. 

We train our video autoencoder and diffusion model on an internal dataset to encode and generate \(256 \times 256\) resolution videos with 9 frames. 
The training and architectural setup closely follow the details outlined in~\Cref{subsubsec:training}. We test on the DAVIS-7 and UCF-7 datasets, as proposed in VIDIM~\cite{jain2024vidim}. These datasets consist of 400 videos, each containing 9 frames, and feature scenes with significant and often ambiguous motion.

~\Cref{table:interpolation} shows the evaluations using reconstruction-based metrics such as PSNR, SSIM, and LPIPS. Although these metrics are commonly used, they can penalize alternative, yet plausible, interpolations. To address this limitation, we also report FVD on the entire video,
providing a more holistic evaluation of interpolation quality. 
We observe that our method performs comparably to VIDIM on the reconstruction metrics. Notably, unlike VIDIM—a diffusion-based baseline requiring a two-stage process with an initial base model followed by a super-resolution step—our model achieves \(256 \times 256\) resolution video generation in a single stage, making it both simpler and more efficient. 
We present qualitative results on two DAVIS scenes in~\Cref{fig:interpolation} to illustrate the effectiveness of our approach in video interpolation. Our method demonstrates comparable quality to the state-of-the-art VIDIM model while producing noticeably sharper details than other methods. These results emphasize the strength of our factorized representation in preserving fine textures and achieving high-fidelity frame generation in complex scenes. 

\subsection{Ablation studies}

To validate our design choices, we conduct ablations on both the factorization and combine methods across the class-conditional and frame prediction tasks, reporting FVD and Inception scores (IS) for the former, and FVD for the latter. We omit IS for frame prediction as it primarily evaluates classifiability and diversity making it unsuitable for the prediction task. See the appendix for additional ablation experiments.

\subsubsection{Factorization}
We explore two variations of the factorization operation (\cref{sec:factorization}). The first approach applies mean pooling (MP) along \(\Lambda_h\), \(\Lambda_w\), and \(\Lambda_t\), effectively reducing dimensionality while preserving essential features. The second employs a learned linear projection (LP) that maps the channel dimesion to \(1\) along the targeted axis. Both methods perform comparably for frame prediction (\cref{table:ablate_factorization}, K600). However, for class-conditional generation (\cref{table:ablate_factorization}, UCF-101), MP significantly outperforms LP. We attribute this gap to the autoencoder's limited generalization, as it is trained on K600. While both methods achieve similar reconstruction results on the K600 test set, with FVD scores of 7.8 (MP) and 8.0 (LP), their performance diverges on UCF, with FVD of 29.5 (MP) versus 37.1 (LP).

\begin{table}[t!]
\centering
\resizebox{0.9\columnwidth}{!}{
\begin{tabular}{@{}llcc@{}}
\toprule \toprule
                         & Factorization Method & FVD$\downarrow$ & Inception$\uparrow$ \\ \midrule
\multirow{2}{*}{UCF-101} & Mean Pooling         & 38  & 91.13     \\
                         & Linear Projection    & 50  & 89.80     \\ \midrule
\multirow{2}{*}{K600}    & Mean Pooling         & 8.6 &   -       \\
                         & Linear Projection    & 8.9 &   -     \\ \bottomrule \bottomrule
\end{tabular}
}
\caption{\textbf{Factorization method.}
We contrast \textit{mean pooling} with \textit{linear projection} for factorizing the volumetric latents.
}
\label{table:ablate_factorization}
\end{table}

\subsubsection{Combine}
We also assess different choices for the combine operation (\cref{sec:recomposition}). Specifically, we test two approaches, \textit{concatenation} defined as \( \mathbf{V}(x, y, t) = [ \mathbf{f}^{1}_{xy} \lvert\lvert \mathbf{f}^{2}_{xy} \lvert \lvert \mathbf{f}_{yt} \lvert \lvert \mathbf{f}^{1}_{xt} ] \), and \textit{summation} given by \( \mathbf{V}(x, y, t) = \mathbf{f}^{1}_{xy} + \mathbf{f}^{2}_{xy} + \mathbf{f}_{yt} + \mathbf{f}^{1}_{xt} \).
\Cref{table:ablate_combine} shows that concatenation yields better performance across both the class-conditional generation and frame prediction task. This improvement likely stems from its ability to retain more distinct feature information from each plane.

\begin{table}[t!]
\centering
\resizebox{0.8\columnwidth}{!}{
\begin{tabular}{@{}llcc@{}}
\toprule \toprule
                         & Combine Method & FVD$\downarrow$ & Inception \\ \midrule
\multirow{2}{*}{UCF-101} & Concat         & 38  & 91.13     \\
                         & Sum            & 45  & 90.76     \\ \midrule
\multirow{2}{*}{K600}    & Concat         & 8.6 &   -        \\
                         & Sum            & 27  &   -       \\ \bottomrule \bottomrule
\end{tabular}
}
\caption{\textbf{Combine method.} 
We contrast \textit{concatenation} with \textit{summation} for recomposing the volume from the factorized latents.
}
\label{table:ablate_combine}
\end{table}

\section{Conclusion}
In this work, we introduced a factorized latent representation that encodes videos into a four-plane structure, paving the way for more efficient representation of spatiotemporal signals.
Coupled with transformer-based diffusion models, our approach enables up to \(2\times\) speedup in training and inference over models operating directly on volumetric latent features—without compromising performance. Our experiments validate that this representation achieves results on par with the previous state-of-the-art across diverse tasks, including class-conditional generation, video extrapolation and interpolation. 
This work presents a simple and effective way to improve the efficiency of models that work with volumetric latent spaces.


%% file: sec/X_suppl.tex
\clearpage
\setcounter{page}{1}
\appendix



\section{Frames vs Reconstruction Quality}
We evaluate the performance of the four-plane factorized representation as the number of video frames increases. Specifically, we test videos with 17, 21, and 25 frames, and report reconstruction performance in~\Cref{table:longer}. We observe that increasing the number of frames from 17 to 25 has only a minor impact on reconstruction quality.

\begin{table}[h!]
\centering
\begin{tabular}{@{}lccc@{}}
\toprule \toprule
  Frames        & PSNR$\uparrow$   & SSIM$\uparrow$  & LPIPS$\downarrow$ \\ \midrule
 17        & 27.11  & 0.82  & 0.051     \\
 21        & 26.95  & 0.82  & 0.051     \\
 25        & 26.51  & 0.81  & 0.052     \\

                     \bottomrule \bottomrule
\end{tabular}
\caption{\textbf{Video reconstruction.} We report reconstruction metric of 4Plane tokenizer for various video lengths.}
\label{table:longer}
\end{table}

\section{Longer Video Generation}
To demonstrate our model's performance on longer sequences, we extend the class-conditional generation experiments to \( 36 \) and \( 56 \) frames, with results presented in~\cref{table:longer_seq}. For \( 36 \) frames, our method achieves comparable performance to WALT while running \( 5\times \) faster. In the \( 56 \)-frame setup, WALT exceeds memory limits due to the increased sequence length, whereas our approach remains efficient.
\begin{table}[h!]
\centering
\begin{tabular}{@{}lcc@{}}
\toprule \toprule
       & \multicolumn{2}{c}{FVD (ms/step)} \\ \midrule
       & 36 frame         & 56 frames      \\ \midrule
WALT   & 164 (1350)       & OOM            \\
\fpshort~(Ours) & 171 (270)        & 564 (378)      \\ \bottomrule \bottomrule
\end{tabular}
\caption{\textbf{Longer Video Generataion.} We report the FVD and training time per step (ms / step) for videos with 36 and 56 frame.}
\label{table:longer_seq}
\end{table}

\section{Timing details}

\begin{figure*}[t!]
\centering
\includegraphics[width=\linewidth]{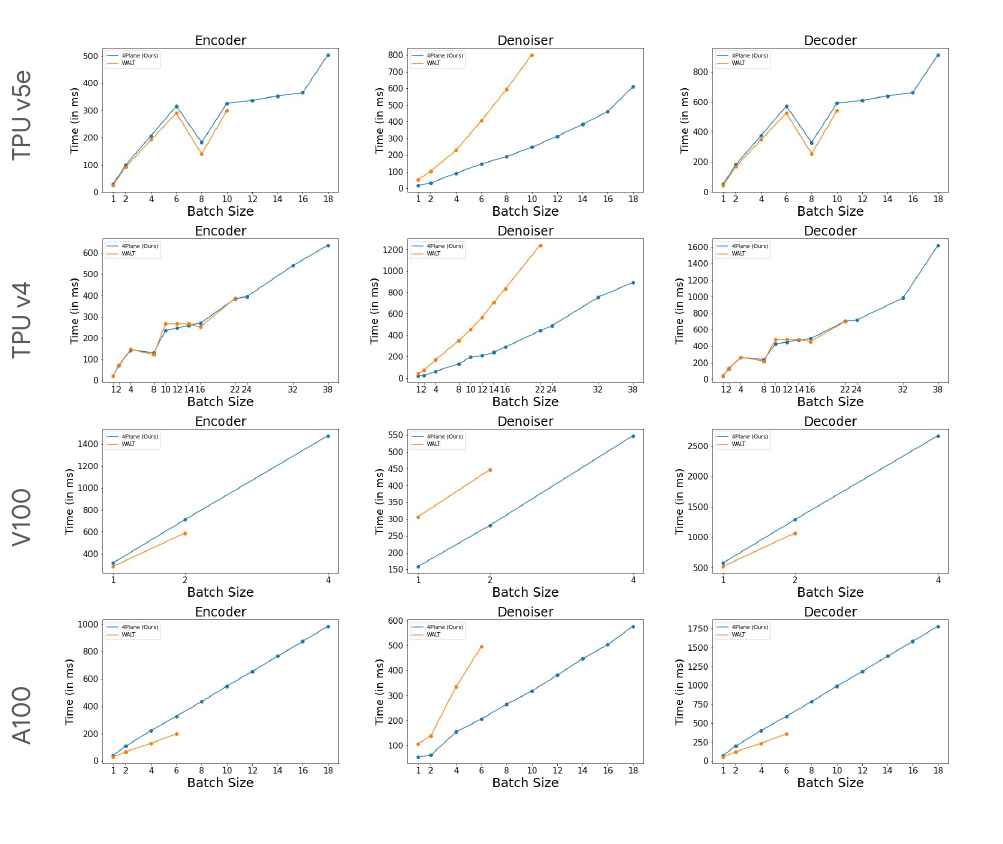}
\caption{\textbf{Timing Breakdown.} Execution times for the encoder, denoiser, and decoder are reported across varying batch sizes on TPU architectures (v5e and v4) in Rows 1 and 2, and GPU architectures (V100 and A100) in Rows 3 and 4. The comparison includes timings for factorized latents (blue) and volumetric latents (orange), measured up to the maximum batch size supported without running out of memory (OOM) for each configuration. The timings are reported for a single step measured during training.}
\label{fig:timing_vlp}
\end{figure*}

A detailed timing breakdown for various components of the model during training, measured with different batch sizes on TPU v5e, TPU v4, V100, and A100 devices in the class-conditional generation setting, is provided in~\Cref{fig:timing_vlp}. These timings were obtained using a model with 214M parameters, alternating between our factorized latent representation and the volumetric latent baseline. For each plot, timings are reported up to the maximum batch size supported on each device. The timings reported in~\Cref{subsec:analysis} correspond to a model trained on a $4 \times 8$ TPU v5e architecture with a batch size of $256$. These measurements approximately align with the timings for a batch size of $8$ shown in row 1 of~\Cref{fig:timing_vlp}.

Across all devices, our model supports larger batch sizes due to its reduced memory requirements. For instance, on TPU v5e (\Cref{fig:timing_vlp}), our model accommodates a batch size of \(18\), whereas the baseline is limited to \(10\). 

The decoder network incurs slightly higher execution time because it contains nearly twice the parameters of the encoder. Although the encoder and decoder in our model are marginally slower than the baseline autoencoder due to the additional factorization and recomposition operations, these operations are executed only once, compared to the denoiser network which is run for \(50\) steps during inference, keeping their overall impact minimal.

During inference on a \(128 \times 128\) video with 17 frames, the 4Plane representation takes 0.17 seconds per video, while the volumetric representation takes 0.40 seconds per video, measured on a \(2 \times 2\) TPU v5e.

\section{Triplane Ablation}

\begin{table}[h!]
\centering
\begin{tabular}{@{}lcc}
\toprule
\toprule
Number of Planes & FVD$\downarrow$   & Inception$\uparrow$ \\ \midrule
Four-plane         & 38    &  91.13     \\
Tri-plane            & 52    & 90.46     \\ 
\bottomrule \bottomrule
\end{tabular}
\caption{\textbf{Ablation: number of planes.} 
We report FVD and Inception scores on the class-conditional task for the UCF-101~\cite{soomro2012ucf101} dataset comparing performance between tri-plane and four plane representation.
}
\label{table:plane_ablate}
\end{table}

To assess the impact of the four-plane representation we perform an experiment where we substitute the four-plane represenation with tri-plane. We report the results in~\Cref{table:plane_ablate}. We only conduct this ablation on the class-conditional task as the frame prediction task cannot be achieved with three plane representation. The superior performance of the four-plane representation over the tri-plane approach can be attributed to its increased capacity for capturing spatial information. By incorporating two spatial planes rather than a single one, the four-plane factorization preserves a more comprehensive set of spatial features, reducing information loss while also providing additional flexibility in its ability to be applied towards frame-conditional tasks in a straightforward manner.

\section{Implementation details}

\subsection{Video autoencoder}
To incorporate image data into the training of the video autoencoder, we adopt an image pretraining strategy commonly employed in prior works~\cite{yu2023magvit,yu2024magvitv2,gupta2023walt}. Specifically, we first train an image autoencoder using 2D convolutional layers. The trained weights are then used to initialize the video autoencoder. Following the approach in MAGVITv2~\cite{yu2024magvitv2}, which shares a similar architecture with our model, we inflate the 2D weights to 3D by initializing the 3D filters to zero and assigning the last slice of the 3D filter to the corresponding 2D filter weights. This method ensures a smooth transition from image-based training to video-based learning, leveraging the pre-trained image representations effectively. For the \( 128 \times 128 \) experiments, the tokenizer consists of \(4\) residual blocks in both the encoder and decoder, with \(2\) temporal downsampling layers and \(3\) spatial downsampling layers. At \( 256 \times 256 \), we increase the capacity to \(5\) residual blocks, maintaining \(2\) temporal downsampling layers while expanding to \(4\) spatial downsampling layers. To trained the autoencoder we use a combination of objectives, including an L2 reconstruction loss, a perceptual loss, and an adversarial loss, to ensure high-quality latent representations that preserve both fine details and overall structure. For the VAE experiments in Section 4.1, we add an additional KL loss with a weight of $10^{-6}$


For class conditional and frame prediction task, we train the tokenizer for \(270,000\) iterations with a batch size of \(256\). The resulting autoencoder achieves a reconstruction performance of \(27.11\) PSNR and \(0.829\) SSIM on videos with \(128 \times 128\) resolution and \(17\) frames. 

For the video interpolation task, the autoencoder is trained for \(450,000\) iterations with the same batch size of \(256\). It achieves a reconstruction PSNR of \(25.58\) and SSIM of \(0.717\) on videos with \(256 \times 256\) resolution and \(9\) temporal frames.

\subsection{Denoiser}
We use the same transformer architecture across all three tasks, following the design and hyperparameters outlined in W.A.L.T.~\cite{gupta2023walt}. 

\begin{itemize}
    \item \textbf{Class-conditional generation}: The denoiser is trained for \(74,000\) iterations with a batch size of \(256\). For the \( 128 \times 128 \) resolution experiments, the input sequence has a length of \(672\), comprising two spatial planes with a resolution of \(16 \times 16\) each and two spatio-temporal planes with a resolution of \(5 \times 16\) each. For the \( 256 \times 256 \) resolution experiments the dimension of the planes and thus the sequence length remains the same due to the additional temporal downsampling.
    \item \textbf{Frame prediction}: The denoiser is trained for \(270,000\) iterations with a batch size of \(256\). The input sequence has a length of \(416\), composed of one spatial plane with a resolution of \(16 \times 16\) and two spatio-temporal planes with a resolution of \(5 \times 16\) each. The conditioning sequence has a length of \(256\), formed by flattening the first spatial plane, which contains information equivalent to the first two latent frames used as conditioning in W.A.L.T.
    \item \textbf{Video interpolation}: The model is trained for \(100,000\) iterations with a batch size of \(256\). The target sequence has a length of \(96\), corresponding to the two spatio-temporal planes, while the conditioning sequence has a length of \(512\).  
\end{itemize}

\subsection{Diffusion}
During training, we adopt a scaled linear noise schedule~\cite{rombach2021high} with \(\beta_0 = 0.0001\) and \(\beta_T = 0.002\), utilizing a DDPM sampler~\cite{ho2020denoising} for the forward diffusion process. During inference, we switch to a DDIM sampler~\cite{song2020denoising} with \(50\) steps.

\section{Joint Image Video Training}
While we have not experimented with it, our four-plane representation can be applied in the joint image-video training setting using the following strategy. When encoding a single image with our 4Plane encoder, the output consists of two identical spatial planes and two spatio-temporal vectors. One of the redundant spatial planes can be discarded, allowing us to use the remaining spatial plane along with the spatio-temporal vectors to train the denoiser network. This results in a slight increase in sequence length compared to the volumetric baseline. For example, given a latent grid of size \(1 \times 16 \times 16\), the sequence length increases from 256 (volumetric) to 288 (4Plane), due to the inclusion of two additional spatio-temporal vectors of size 16 each. This strategy allows for using the same autoencoder and denoiser network for images and videos.